\documentclass{article}

\newcommand{\jp}[1]{}
\newcommand{\en}[1]{#1}

\newcommand{\mainpaper}[1]{#1}
\newcommand{\supplementary}[1]{#1}

\usepackage[final]{neurips_2019}

\usepackage[utf8]{inputenc} 
\usepackage[T1]{fontenc}    
\usepackage{hyperref}       
\usepackage{url}            
\usepackage{booktabs}       
\usepackage{amsfonts}       
\usepackage{nicefrac}       
\usepackage{microtype}      

\usepackage[pdftex]{graphicx}
\usepackage{amsmath, amssymb}

\newcommand{\z}{^{(0)}}
\renewcommand{\o}{^{(1)}}
\newcommand{\w}{^{(2)}}
\newcommand{\h}{^{(3)}}

\newcommand{\figref}[1]{\figurename \ \ref{#1}}
\newcommand{\vect}[1]{{\boldsymbol{#1}}}
\newcommand{\gyouretu}[4]{\left(
    \begin{array}{cc}
      {#1} & {#2} \\
      {#3} & {#4}
    \end{array}
  \right)}
\newcommand{\gyouretsu}[9]{\left(
    \begin{array}{ccc}
      {#1} & {#2} & {#3} \\
      {#4} & {#5} & {#6} \\
      {#7} & {#8} & {#9}
    \end{array}
  \right)}
\newcommand{\gyouretuthree}[3]{\left(
\begin{array}{cc}
  {#1} & {#2} \\
  {} & {#3}
\end{array}
\right)}
\newcommand{\gyouretusix}[6]{\left(
    \begin{array}{ccc}
      {#1} & {#2} & {#3} \\
      {} & {*} & {#5} \\
      {} & {} & {#6}
    \end{array}
  \right)}  
\newcommand{\gyouretuseven}[7]{\scalebox{0.8}{$\left(
    \begin{array}{cccc}
      {Q_{ii}} & {Q_{ij}} & {#1} & {#2} \\
      {} & {Q_{jj}} & {#3} & {#4} \\
      {} & {} & {#5} & {#6} \\
      {} & {} & {} & {#7}
    \end{array}
  \right)$}}
\newcommand{\gyouretuseveno}[7]{\scalebox{0.8}{$\left(
    \begin{array}{cccc}
      {Q\o_{ii}} & {Q\o_{ij}} & {#1} & {#2} \\
      {} & {Q\o_{jj}} & {#3} & {#4} \\
      {} & {} & {#5} & {#6} \\
      {} & {} & {} & {#7}
    \end{array}
  \right)$}}

\newcommand{\titlestr}{Data-Dependence of Plateau Phenomenon in Learning with Neural Network --- Statistical Mechanical Analysis}

\title{\titlestr}

\author{%
  Yuki Yoshida \qquad\qquad Masato Okada \\
  Department of Complexity Science and Engineering, Graduate School of Frontier Sciences, \\ The University of Tokyo \\
  5-1-5 Kashiwanoha, Kashiwa, Chiba 277-8561, Japan \\
  \texttt{\{yoshida@mns, okada@edu\}.k.u-tokyo.ac.jp} \\
}

\begin{document}
\mainpaper{

\maketitle

\begin{abstract}
\en{The plateau phenomenon, wherein the loss value stops decreasing during the process of learning, has been reported by various researchers. The phenomenon is actively inspected in the 1990s and found to be due to the fundamental hierarchical structure of neural network models. Then the phenomenon has been thought as inevitable. However, the phenomenon seldom occurs in the context of recent deep learning. There is a gap between theory and reality. In this paper, using statistical mechanical formulation, we clarified the relationship between the plateau phenomenon and the statistical property of the data learned. It is shown that the data whose covariance has small and dispersed eigenvalues tend to make the plateau phenomenon inconspicuous. 
}
\jp{ニューラルネットワークの学習中に誤差の減少が停滞する「プラトー現象」が問題となっていた．プラトー現象は1990年代に様々な文献で報告および研究され，ニューラルネットワークモデルの階層性に由来する理論上不可避な現象と考えられていた．しかし一方で，近年の深層学習においてはプラトー現象が問題となることはほとんどなく，理論と現実に乖離がある．本研究では，データの分布の統計性とプラトー現象の関係性を明らかにする．その結果，入力の実効的なノルムがプラトーの性質を決定づけることが明らかとなった．}
\end{abstract}

\section{Introduction}
\subsection{Plateau Phenomenon}
\en{
Deep learning, and neural network as its essential component, has come to be applied to various fields. However, these still remain unclear in various points theoretically. The plateau phenomenon is one of them. 
In the learning process of neural networks, their weight parameters are updated iteratively so that the loss decreases.
However, in some settings the loss does not decrease simply, but its decreasing speed slows down significantly partway through learning, and then it speeds up again after a long period of time. This is called as ``plateau phenomenon''. Since 1990s, this phenomena have been reported to occur in various practical learning situations (see \figref{fig_first_experiment} (a) and \citet{park2000adaptive, fukumizuamari2000}) .
As a fundamental cause of this phenomenon, it has been pointed out by a number of researchers that the intrinsic symmetry of neural network models brings singularity to the metric in the parameter space which then gives rise to special attractors whose regions of attraction have nonzero measure, called as Milnor attractor (defined by \citet{milnor1985}; see also Figure 5 in  \cite{fukumizuamari2000} for a schematic diagram of the attractor).
}
\jp{深層学習，およびその構成要素であるニューラルネットワークは，様々な分野に応用されるようになった．しかしながら，その理論的側面には明らかでない点も多い．
そのひとつに，プラトー現象がある．
プラトー現象とは，ニューラルネットの学習中に誤差の減少が長時間にわたり停滞する現象である．これは，いわゆる悪い局所解とは異なり，プラトー現象が終了すると誤差が再び減少を再開する．
プラトー現象は，1990年代頃から様々な学習ケースにおいて実際に生じることが報告されてきた（文献〜〜〜〜，および\figref{fig_first_experiment} (a) を参照されたい）．
プラトー現象の根本原因として，ニューラルネットワークのモデルに内在する対称性が，パラメータ空間の計量に特異性をもたらし，それが，非ゼロの測度の吸引領域をもつアトラクタ，いわゆるミルナーアトラクタを生み出すことが，様々な研究により指摘されてきた．}

\subsection{Who moved the plateau phenomenon?} 
\en{However, the plateau phenomenon seldom occurs in
recent practical use of neural networks
(see \figref{fig_first_experiment} (b) for example).}
\jp{一方で，近年の深層学習において，プラトー現象が問題となることは，様々な領域においてほとんど報告されていない（例えば，\figref{fig_first_experiment} (b) を参照されたい）．}

\en{In this research, we rethink the plateau phenomenon, and discuss which situations are likely to cause the phenomenon. First we introduce the student-teacher model of two-layered networks as an ideal system. Next, we reduce the learning dynamics of the student-teacher model to a small-dimensional order parameter system by using statistical mechanical formulation, under the assumption that the input dimension is sufficiently large. Through analyzing the order parameter system, we can discuss how the macroscopic learning dynamics depends on the statistics of input data. Our main contribution is the following:
\begin{itemize}
    \item Under the statistical mechanical formulation of learning in the two-layered perceptron, we showed that macroscopic equations can be derived even when the statistical properties of the input are generalized. In other words, we extended the result of \citet{saadsolla1995} and \citet{rieglerbiehl1995}.
    \item By analyzing the macroscopic system we derived, we showed that the dynamics of learning depends only on the eigenvalue distribution of the covariance matrix of the input data.
    \item We clarified the relationship between the input data statistics and plateau phenomenon. In particular, it is shown that the data whose covariance matrix has small and disparsed eigenvalues tend to make the phenomenon inconspicuous, by numerically analyzing the macroscopic system. 
\end{itemize}
}
\jp{本研究では，プラトー現象を再考し，プラトー現象がどのような状況で生じるかを論じる．
理想的な系として，まず三層ネットワークの生徒教師モデルを導入する．
次に，入力次元数が十分に大きいという仮定のもとで，統計力学的定式化を用いて，
生徒教師モデルの学習ダイナミクスを少変数系に縮約する．
オーダパラメータ系を解析することで，プラトー現象を含む学習中のマクロな挙動が，
入力データの持つ統計性にどのように依存するかを議論する．主な貢献は以下の通りである：
\begin{itemize}
    \item 三層パーセプトロンの学習の統計力学的定式化において，入力の統計性が一般化された場合においても巨視的方程式が導出可能であることを示した．すなわち，Saad らの結果を拡張した．
    \item 導出した巨視的方程式を解析することにより，学習係数が小さな時には，学習ダイナミクスが入力の共分散行列の固有値分布にのみ依存していることを示した．
    \item 入力の共分散行列の固有値分布の平均値が大きなほどプラトーが短縮すること，および，平均が同じ場合にはその分散が大きなほどプラトーが短縮することを，巨視的方程式を数値的に解析することにより明らかにした．
\end{itemize}
}

\subsection{Related works}
\en{
The statistical mechanical approach used in this research is firstly developed by \citet{saadsolla1995}. The method reduces high-dimensional learning dynamics of nonlinear neural networks to low-dimensional system of order parameters. They derived the macroscopic behavior of learning dynamics in two-layered soft-committee machine and by analyzing it they point out the existence of plateau phenomenon. Nowadays the statistical mechanical method is applied to analyze recent techniques (\citet{hara2016analysis}, \citet{yoshida2017statistical}, \citet{takagi2019impact} and \citet{straat2019line}), and generalization performance in over-parameterized setting (\citet{goldt2019dynamics}) and environment with conceptual drift (\citet{straat2018statistical}). However, it is unknown that how the property of input dataset itself can affect the learning dynamics, including plateaus.
}\jp{
本論文で用いる統計力学的手法は，\citet{saadsolla1995} らによって始められたものである．
この手法は，非線形ニューラルネットの高次元の学習ダイナミクスを低次元のオーダパラメータ系に縮約するものである．彼らは，三層ソフトコミッティマシンの学習ダイナミクスに関して，その巨視的な挙動を導出し，
それを数値的に解析することにより，学習中にプラトー現象が生じることを指摘している．近年では，この手法は Dropout の解析(Hara)，Layer Normalization の解析(Takagi)，Weight Normalization の解析(Yoshida)，ReLU ネットの解析(Biehl)，Over-param での汎化誤差の解析 (Goldt) に用いられている．また concept drift を伴う continual な環境での学習ダイナミクスが Biehl らによって解析されている．
しかしながら，学習ダイナミクスに対して，入力データセットそのものの統計性がどのように影響を与えるかに関しては，明らかとなっていない．
}

\en{
Plateau phenomenon and singularity in loss landscape as its main cause have been studied by \citet{fukumizuamari2000}, \citet{wei2008dynamics}, \citet{cousseau2008dynamics} and \citet{guo2018influence}. On the other hand, recent several works suggest that plateau and singularity can be mitigated in some settings. \citet{orhan2017skip} shows that skip connections eliminate the singularity. Another work by \citet{yoshida2019statistical} points out that output dimensionality affects the plateau phenomenon, in that multiple output units alleviate the plateau phenomenon. However, the number of output elements does not fully determine the presence or absence of plateaus, nor does the use of skip connections. The statistical property of data just can affect the learning dynamics dramatically; for example, see \figref{fig_stwvid} for learning curves with using different datasets and same network architecture. We focus on what kind of statistical property of the data brings plateau phenomenon.
}
\jp{
プラトーおよびその原因である singularity については，Fukumizu, Cousseau, Wei など．近年 Wei が singularity の influence area が大きいことを指摘している．一方，プラトーが生じなくなるケースを指摘する研究もある．
近年， \citet{orhan2017skip} らは，スキップ結合が プラトー現象の原因である singularity を消失させることを示した．
また， \citet{yoshida2019statistical} らは，プラトー現象を引き起こす要因の一つとして出力素子数を指摘している．彼らの研究では，出力の次元数が複数の場合にはプラトー現象が起こりづらくなることが指摘されている．
しかしながら，出力素子の個数やスキップ結合の存在だけがプラトーの有無を決定しているわけではない．データの統計性だけでも，学習ダイナミクスは劇的に変わる場合がある：
例えば \figref{fig_stwvid} は，同一のネットワーク構造に異なるデータセットを与えた場合の学習曲線である．本研究では，いかなる入力データの統計性がプラトー現象を生じさせるか，という観点を論じる．
}

\begin{figure}[htbp]
\begin{center}
\includegraphics[width=0.7\linewidth]{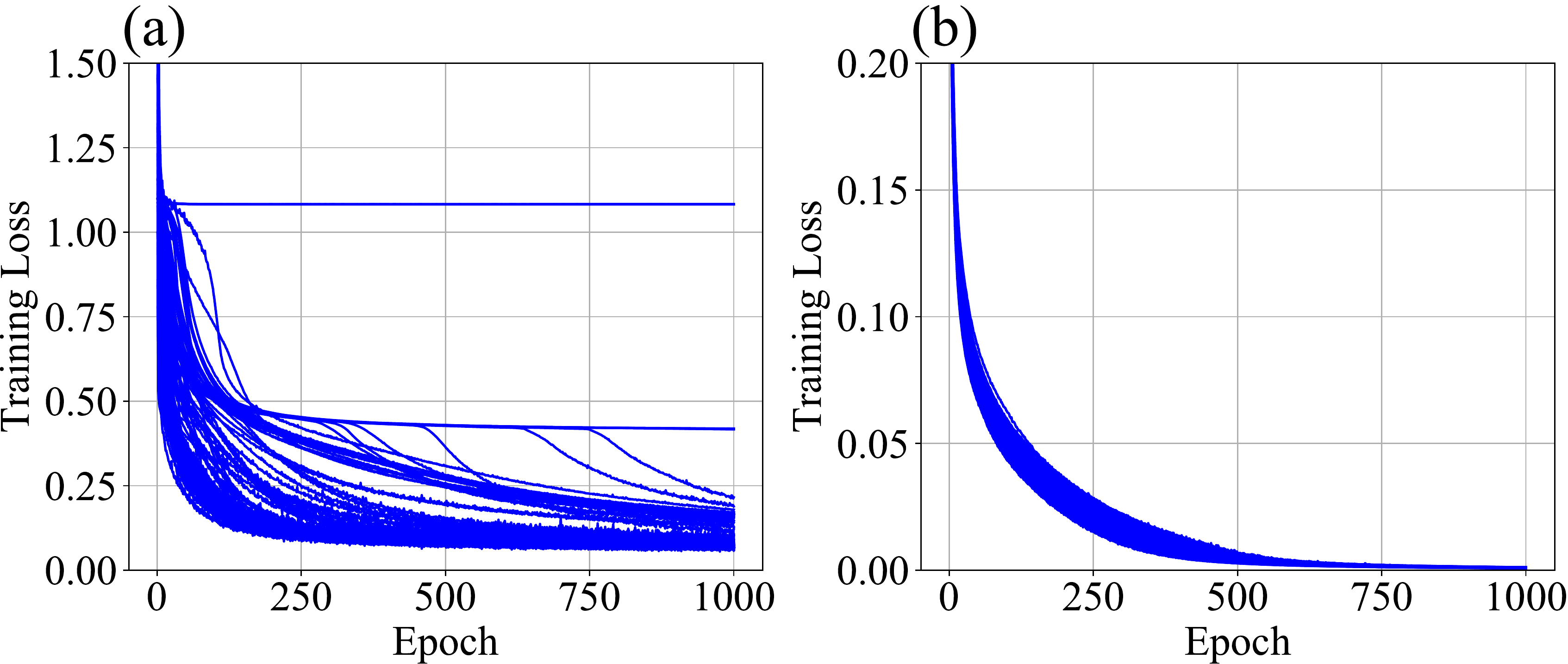}
\caption{
\en{
 (a) Training loss curves when two-layer perceptron with $4$-$4$-$3$ units and ReLU activation learns IRIS dataset. 
 (b) Training loss curve when two-layer perceptron with $784$-$20$-$10$ units and ReLU activation learns MNIST dataset.
 For both (a) and (b), results of 100 trials with random initialization are overlaid. Minibatch size of 10 and vanilla SGD (learning rate: 0.01) are used.
}
\jp{
 (a) IRIS データセットを，素子数 $4$-$4$-$3$ の三層パーセプトロン（ReLU 活性化）に学習させた際の loss curve．ランダムな異なる 100 通りの重み初期値に対する試行結果を重ね書きしている．
 (b) MNIST データセットを，素子数 $784$-$20$-$10$ の三層パーセプトロン（ReLU 活性化）に学習させた際の loss curve．ランダムな異なる 20 通りの重み初期値に対する試行結果を重ね書きしている．
 (a)(b) ともに勾配法は vanilla SGD を用いており，ミニバッチサイズは 10 である．
}
}
\label{fig_first_experiment}
\end{center}
\end{figure}

\begin{figure}[htbp]
\begin{center}
\includegraphics[width=0.95\linewidth]{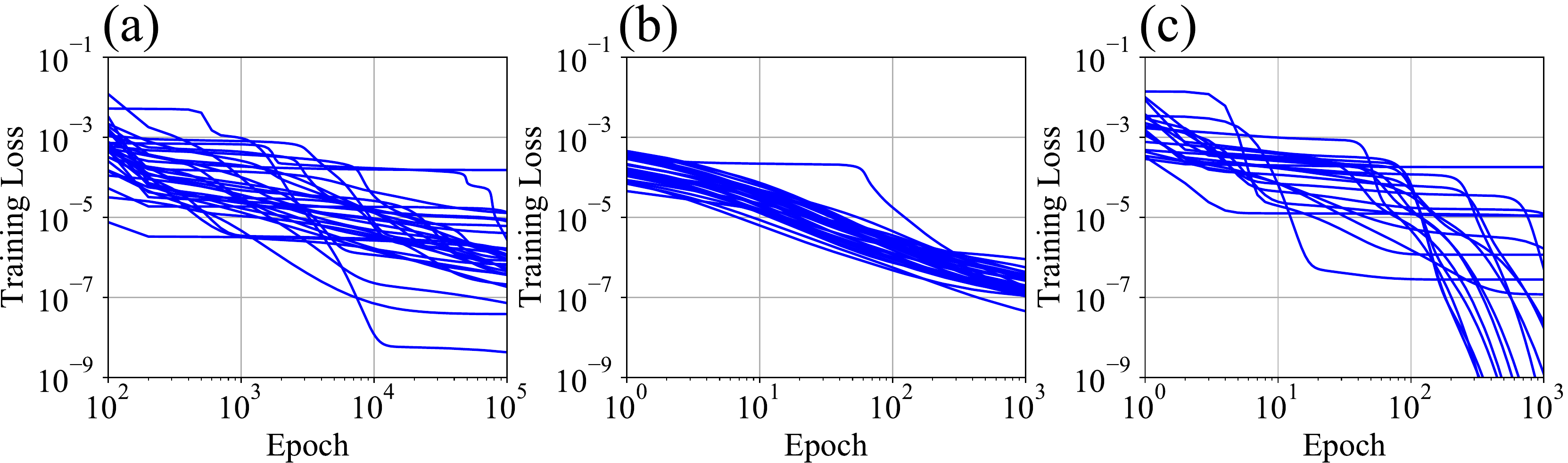}
\caption{
\en{
 Loss curves yielded by student-teacher learning with two-layer perceptron which has 2 hidden units, 1 output unit and sigmoid activation, and with (a) IRIS dataset, (b) MNIST dataset, (c) a dataset in $\mathbb{R}^{60000\times 784}$ drawn from standard normal distribution, as input distribution $p(\vect{\xi})$.
 In every subfigure, results for 20 trials with random initialization are overlaid. Vanilla SGD (learning rate: (a)(b) 0.005, (c) 0.001) and minibatch size of 1 are used for all three settings.
}
\jp{
 (a) IRIS データセット，(b) MNIST データセット，
 (c) 標準ガウス分布 を入力分布 $p(\vect{\xi})$ に用いて，三層パーセプトロン（sigmoid 活性化）の生徒教師学習を行った場合の loss curve. それぞれ，20 通りのランダムな重み初期値に対する試行結果を重ね書きしている．(a)(b)(c) いずれも中間素子数は 2，出力素子数は 1 である．勾配法は SGD，ミニバッチサイズは 1 である．
}
}
\label{fig_stwvid}
\end{center}
\end{figure}

\section{Formulation}
\subsection{Student-Teacher Model}
\en{
We consider a two-layer perceptron which has $N$ input units, $K$ hidden units and $1$ output unit. We denote the input to the network by $\vect{\xi} \in \mathbb{R}^N$. Then the output can be written as $s = \sum_{i=1}^K w_i g(\vect{J}_i\cdot\vect{\xi}) \in \mathbb{R}$, where $g$ is an activation function.
}
\jp{
入力次元 $N$，中間次元 $K$，出力次元 1 の三層パーセプトロンを考える．
このネットワークの入力を $\vect{\xi} \in \mathbb{R}^N$ と書くならば，
出力は $s = \sum_{i=1}^K w_i g(\vect{J}_i\cdot\vect{\xi}) \in \mathbb{R}$ と表される．ただし $g$ は活性化関数を表す．
}

\en{
We consider the situation that the network learns data generated by another network, called ``teacher network'', which has fixed weights. Specifically, we consider two-layer perceptron that outputs $t = \sum_{n=1}^M v_n g(\vect{B}_n\cdot\vect{\xi}) \in \mathbb{R}$ for input $\vect{\xi}$ as the teacher network. The generated data $(\vect{\xi}, t)$ is then fed to the student network stated above and learned by it in the on-line manner (see \figref{fig_nets}). We assume that the input $\vect{\xi}$ is drawn from some distribution $p(\vect{\xi})$ every time independently. 
We adopt vanilla stochastic gradient descent (SGD) algorithm for learning.
We assume the squared loss function $\varepsilon = \frac12 (s - t)^2$, which is most commonly used for regression.
}
\jp{
本稿では，実データを用いた学習を考える代わりに，ある「教師ネットワーク」が
生成するデータを学習する状況を考える．
より具体的には，入力 $\vect{\xi}$ に対して
$t = \sum_{n=1}^M v_n g(\vect{B}_n\cdot\vect{\xi}) \in \mathbb{R}$ を出力する三層パーセプトロン（教師ネットワーク）を考え，これが生成するデータ
$(\vect{\xi}, t)$ を上述の生徒ネットワークが逐次的にオンライン学習する状況を考える
（\figref{fig_nets}(a)）．ただし，入力 $\vect{\xi}$ は確率分布 $p(\vect{\xi})$ から
毎ステップ独立に生成されるものとする．
また，学習のアルゴリズムとしては，本稿では標準的な勾配法を考える．損失関数としては，
二乗誤差 $\mathcal{L}(s, t) := \frac12(s - t)^2$ を考える．
}

\subsection{Statistical Mechanical Formulation}
\en{
In order to capture the learning dynamics of nonlinear neural networks described in the previous subsection macroscopically, we introduce the statistical mechanical formulation in this subsection.
}
\jp{
前小節の設定の下での非線形ニューラルネットの学習ダイナミクスを，粗視的に捉えるため，本小節では統計力学的定式化を導入する．
}


\en{
Let $x_i := \vect{J}_i\cdot\vect{\xi}$ ($1 \leq i \leq K$) and $y_n := \vect{B}_n\cdot\vect{\xi}$ ($1 \leq n \leq M$). Then
\begin{align*}
&\quad (x_1, \ldots, x_K, y_1, \ldots, y_M) 
\sim \mathcal{N} \left( 0, [\vect{J}_1, \ldots, \vect{J}_K, \vect{B}_1, \ldots, \vect{B}_M]^T \Sigma [\vect{J}_1, \ldots, \vect{J}_K, \vect{B}_1, \ldots, \vect{B}_M]\right)
\end{align*}
holds with $N \to \infty$ by generalized central limit theorem, provided that the input distribution $p(\vect{\xi})$ has zero mean and finite covariance matrix $\Sigma$.
}\jp{
まず，入力 $\vect{\xi}$ は中心 $\vect{0}$, 共分散行列 $\Sigma$ の多変量ガウス分布
$\mathcal{N}(\vect{0}, \Sigma)$ に従うものとする．このとき，
\begin{align*}
(x_i, x_j, x_k) \sim 
\mathcal{N}
\left(
	\vect{0},
	\gyouretsu{\vect{J}_i^T\Sigma\vect{J}_i}{\vect{J}_i^T\Sigma\vect{J}_j}{\vect{J}_i^T\Sigma\vect{J}_k}{\vect{J}_j^T\Sigma\vect{J}_i}{\vect{J}_j^T\Sigma\vect{J}_j}{\vect{J}_j^T\Sigma\vect{J}_k}{\vect{J}_k^T\Sigma\vect{J}_i}{\vect{J}_k^T\Sigma\vect{J}_j}{\vect{J}_k^T\Sigma\vect{J}_k}
\right)
\end{align*}
が成立する．
}

\en{
Next, let us introduce order parameters as following:
$Q_{ij} := \vect{J}_i^T\Sigma\vect{J}_j = \langle x_ix_j \rangle$, \
$R_{in} := \vect{J}_i^T\Sigma\vect{B}_n = \langle x_iy_n \rangle$, \ 
$T_{nm} := \vect{B}_n^T\Sigma\vect{B}_m = \langle y_ny_m \rangle$
and
$D_{ij} := w_i w_j$, \
$E_{in} := w_i v_n$, \
$F_{nm} := v_n v_m$.
Then
\begin{align*}
(x_1, \ldots, x_K, y_1, \ldots, y_M) \sim
\mathcal{N}(\vect{0}, \gyouretu{Q}{R}{R^T}{T}).
\end{align*}
}
\jp{
次に，オーダパラメータを以下のように導入する：
$Q_{ij} := \vect{J}_i^T\Sigma\vect{J}_j = \langle x_ix_j \rangle$, \
$R_{in} := \vect{J}_i^T\Sigma\vect{B}_n = \langle x_iy_n \rangle$, \ 
$T_{nm} := \vect{B}_n^T\Sigma\vect{B}_m = \langle y_ny_m \rangle$
and
$D_{ij} := \vect{w}_i^T\vect{w}_j$, \
$E_{in} := \vect{w}_i^T\vect{v}_n$, \
$F_{nm} := \vect{v}_n^T\vect{v}_m$.
}

The parameters $Q_{ij}$, $R_{in}$, $T_{nm}$, $D_{ij}$, $E_{in}$, and $F_{nm}$ introduced above capture
the state of the system macroscopically; therefore they are called as ``order parameters.'' The first three represent the state of the first layers of the two networks (student and teacher), and
the latter three represent their second layers' state.
$Q$ describes the statistics of the student's first layer and $T$ represents that of the teacher's first layer.
$R$ is related to similarity between the student and teacher's first layer.
$D, E, F$ is the second layers' counterpart of $Q, R, T$.
The values of $Q_{ij}$, $R_{in}$, $D_{ij}$, and $E_{in}$ change during learning;
their dynamics are what to be determined, from the dynamics of microscopic variables, i.e. connection weights. In contrast, $T_{nm}$ and $F_{nm}$ are constant during learning.

\begin{figure}[htbp]
\begin{center}
\includegraphics[width=0.75\linewidth]{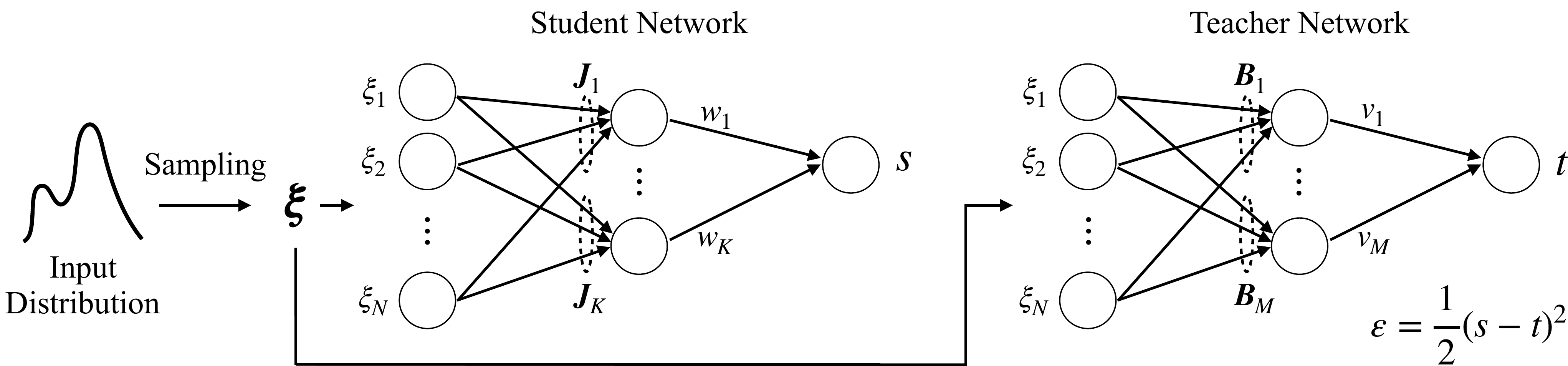}
\caption{
Overview of student-teacher model formulation.
}
\label{fig_nets}
\end{center}
\end{figure}

\subsubsection{Higher-order order parameters}
\en{
The important difference between our situation and that of \citet{saadsolla1995} is the covariance matrix $\Sigma$ of the input $\vect{\xi}$ is not necessarily equal to identity. This makes the matter complicated, since higher-order terms $\Sigma^e$ ($e = 1, 2, \ldots$) appear inevitably in the learning dynamics of order parameters. In order to deal with these, here we define some higher-order version of order parameters.

Let us define higher-order order parameters $Q_{ij}^{(e)}$, $R_{in}^{(e)}$ and $T_{nm}^{(e)}$ for $e = 0, 1, 2, \ldots$, as
$
Q_{ij}^{(e)} := \vect{J}_i^T\Sigma^e\vect{J}_j, \quad
R_{in}^{(e)} := \vect{J}_i^T\Sigma^e\vect{B}_n, \quad \text{and} \quad
T_{nm}^{(e)} := \vect{B}_n^T\Sigma^e\vect{B}_m.
$
Note that they are identical to $Q_{ij}$, $R_{in}$ and  $T_{nm}$ in the case of $e = 1$.
Also we define higher-order version of $x_i$ and $y_n$, namely $x^{(e)}_i$ and $y^{(e)}_n$, as
$
x^{(e)}_i := \vect{\xi}^T \Sigma^e \vect{J}_i,
y^{(e)}_n := \vect{\xi}^T \Sigma^e\vect{B}_n.
$
Note that $x^{(0)}_i = x_i$ and $y^{(0)}_n = y_n$.
}
\jp{
さらに，$e = 0, 1, 2, \ldots$ に対して，高階のオーダパラメータ
$Q_{ij}^{(e)}$, $R_{in}^{(e)}$, $T_{nm}^{(e)}$ を以下により定義する：
\begin{align*}
Q_{ij}^{(e)} &:= \vect{J}_i^T\Sigma^e\vect{J}_j, \\
R_{in}^{(e)} &:= \vect{J}_i^T\Sigma^e\vect{B}_n, \\ 
T_{nm}^{(e)} &:= \vect{B}_n^T\Sigma^e\vect{B}_m.
\end{align*}
なお $e = 1$ の場合，これらはそれぞれ $Q_{ij}$, $R_{in}$, $T_{nm}$ と一致することに注意されたい．
Also we define higher-order version of $x_i$ and $y_n$, namely $x^{(e)}_i$ and $y^{(e)}_n$, as
\begin{align*}
x^{(e)}_i &:= \vect{\xi}^T \Sigma^e \vect{J}_i, \\
y^{(e)}_n &:= \vect{\xi}^T \Sigma^e\vect{B}_n.
\end{align*}
Note that $x^{(0)}_i$ and $y^{(0)}_n$ are identical to $x_i$ and $y_n$ respectively.
}

\section{Derivation of dynamics of order parameters}
\en{
At each iteration of on-line learning, weights of the student network $\vect{J}_i$ and $w_i$ are updated with
\begin{equation}
\begin{aligned}
\Delta \vect{J}_i &= -\frac{\eta}{N} \frac{d\varepsilon}{d\vect{J}_i} = \frac{\eta}{N} [(\vect{t}-\vect{s
})\cdot\vect{w}_i] g'(x_i) \vect{\xi} 
= \frac{\eta}{N} \left[ \left( \sum_{n=1}^M \vect{v}_n g(y_n) - \sum_{j=1}^K \vect{w}_j g(x_j) \right) \cdot\vect{w}_i \right] g'(x_i) \vect{\xi},  \\
\Delta \vect{w}_i &= -\frac{\eta}{N} \frac{d\varepsilon}{d\vect{w}_i} = \frac{\eta}{N} g(x_i) (\vect{t}-\vect{s}) 
= \frac{\eta}{N} g(x_i) \left(\sum_{n=1}^M \vect{v}_n g(y_n) - \sum_{j=1}^K \vect{w}_j g(x_j) \right),
\end{aligned} \label{10}
\end{equation}
in which we set the learning rate as $\eta/N$, so that our macroscopic system is $N$-independent.
}
\jp{
生徒ネットワークがオンライン学習を行う各ステップにおいて，生徒の結合荷重
$\vect{J}_i$ および $w_i$ は，以下の式に従って更新される：
\begin{equation}
\begin{aligned}
\Delta \vect{J}_i &= -\frac{\eta}{N} \frac{d\varepsilon}{d\vect{J}_i} = \frac{\eta}{N} [(\vect{t}-\vect{s
})\cdot\vect{w}_i] g'(x_i) \vect{\xi} \\
&= \frac{\eta}{N} \left[ \left( \sum_{n=1}^M \vect{v}_n g(y_n) - \sum_{j=1}^K \vect{w}_j g(x_j) \right) \cdot\vect{w}_i \right] g'(x_i) \vect{\xi},  \\
\Delta \vect{w}_i &= -\frac{\eta}{N} \frac{d\varepsilon}{d\vect{w}_i} = \frac{\eta}{N} g(x_i) (\vect{t}-\vect{s}) \\
&= \frac{\eta}{N} g(x_i) \left(\sum_{n=1}^M \vect{v}_n g(y_n) - \sum_{j=1}^K \vect{w}_j g(x_j) \right),
\end{aligned} \label{10}
\end{equation}
in which we set the learning rate as $\eta/N$, so that our macroscopic system is $N$-independent.
}

Then, the order parameters $Q_{ij}^{(e)}$ and $R_{in}^{(e)}$ ($e = 0, 1, 2, \ldots$) are updated with
\begin{equation}
\begin{aligned}
\Delta Q_{ij}^{(e)} &= (\vect{J}_i + \Delta\vect{J}_i)^T \Sigma^e (\vect{J}_j + \Delta\vect{J}_j) - \vect{J}_i^T \Sigma^e \vect{J}_j
= \vect{J}_i^T \Sigma^e \Delta \vect{J}_j + \vect{J}_j^T \Sigma^e \Delta \vect{J}_i + \Delta \vect{J}_i^T \Sigma^e \Delta \vect{J}_j \\
&= \frac{\eta}{N} \left[ \sum_{p=1}^M E_{ip} g'(x_i)x_j^{(e)}g(y_p) - \sum_{p=1}^K D_{ip} g'(x_i)x_j^{(e)}g(x_p)\right. \\
&\qquad\qquad +  \left. \sum_{p=1}^M E_{jp} g'(x_j)x_i^{(e)}g(y_p) - \sum_{p=1}^K D_{jp} g'(x_j)x_i^{(e)}g(x_p) \right] \\
&+ \frac{\eta^2}{N^2} \vect{\xi}^T \Sigma^e \vect{\xi} \left[
\sum_{p, q}^{K, K} D_{ip}D_{jq} g'(x_i)g'(x_j)g(x_p)g(x_q) \right.
\left. + \sum_{p, q}^{M, M} E_{ip}E_{jq} g'(x_i)g'(x_j)g(y_p)g(y_q) \right. \\
&\qquad \left. - \sum_{p, q}^{K, M} D_{ip}E_{jq} g'(x_i)g'(x_j)g(x_p)g(y_q) \right.
\left. - \sum_{p, q}^{M, K} E_{ip}D_{jq} g'(x_i)g'(x_j)g(y_p)g(x_q) \right], \\
\Delta R_{in}^{(e)} &= (\vect{J}_i + \Delta\vect{J}_i)^T \Sigma^e \vect{B}_n - \vect{J}_i^T \Sigma^e \vect{B}_n
= \Delta \vect{J}_i^T \Sigma^e \vect{B}_n \\
&= \frac{\eta}{N} \left[ \sum_{p=1}^M E_{ip} g'(x_i)y_n^{(e)}g(y_p) - \sum_{p=1}^K D_{ip} g'(x_i)y_n^{(e)}g(x_p) \right].
\end{aligned}\label{21}
\end{equation}

Since
\begin{align*}
    &\vect{\xi}^T \Sigma^e \vect{\xi} \approx N \mu_{e+1} \qquad
    \text{where}\quad \mu_d := \frac1N \sum_{i=1}^N \lambda_i^d,  \qquad
    \lambda_1, \ldots, \lambda_N : \text{eigenvalues of $\Sigma$}
\end{align*}
and the right hand sides of the difference equations are $O(N^{-1})$,
we can replace these difference equations with differential ones with $N \to \infty$, by taking the expectation over all
input vectors $\vect{\xi}$:
\begin{equation}
\begin{aligned}
\frac{dQ_{ij}^{(e)}}{d\tilde{\alpha}}
&= \eta \left[ \sum_{p=1}^M E_{ip} I_3(x_i, x^{(e)}_j, y_p) - \sum_{p=1}^K D_{ip} I_3(x_i, x^{(e)}_j, x_p) \right. \\
&\qquad\qquad \left. +\sum_{p=1}^M E_{jp} I_3(x_j, x^{(e)}_i, y_p) - \sum_{p=1}^K D_{jp} I_3(x_j, x^{(e)}_i, x_p) \right] \\
&+ \eta^2 \mu_{e+1} \left[
\sum_{p, q}^{K, K} D_{ip}D_{jq} I_4(x_i, x_j, x_p, x_q) \right.
\left. + \sum_{p, q}^{M, M} E_{ip}E_{jq} I_4(x_i, x_j, y_p, y_q) \right. \\
&\qquad \left. - \sum_{p, q}^{K, M} D_{ip}E_{jq} I_4(x_i, x_j, x_p, y_q) \right.
\left. - \sum_{p, q}^{M, K} E_{ip}D_{jq} I_4(x_i, x_j, y_p, x_q) \right] , \\
\frac{dR_{in}^{(e)}}{d\tilde{\alpha}}
&= \eta \left[ \sum_{p=1}^M E_{ip} I_3(x_i, y^{(e)}_n, y_p) - \sum_{p=1}^K D_{ip} I_3(x_i, y^{(e)}_n, x_p) \right]
\end{aligned}\label{5030}\end{equation}
\begin{equation}
\begin{aligned}
\text{where} \quad
I_3(z_1, z_2, z_3) &:= \langle g'(z_1)z_2g(z_3)\rangle \quad\text{and}\quad
I_4(z_1, z_2, z_3, z_4) := \langle g'(z_1)g'(z_2)g(z_3)g(z_4)\rangle.
\end{aligned}\label{40}
\end{equation}
In these equations, $\tilde{\alpha} := \alpha/N$ represents time (normalized number of steps), and the brackets $\langle\cdot\rangle$
represent the expectation when the input $\vect{\xi}$ follows the input distribution $p(\vect{\xi})$.


The differential equations for $D$ and $E$ are obtained in a similar way:
\begin{equation}
\begin{aligned}
\frac{dD_{ij}}{d\tilde{\alpha}}
&= \eta \left[ \sum_{p=1}^M E_{ip} I_2(x_j, y_p) - \sum_{p=1}^K D_{ip} I_2(x_j, x_p) \right.
\left. +  \sum_{p=1}^M E_{jp} I_2(x_i, y_p) - \sum_{p=1}^K D_{jp} I_2(x_i, x_p) \right], \\
\frac{dE_{in}}{d\tilde{\alpha}}
&= \eta \left[ \sum_{p=1}^M F_{pn} I_2(x_i, y_p) - \sum_{p=1}^K E_{pn} I_2(x_i, x_p) \right]
\end{aligned}\label{60}
\end{equation}
\begin{equation}
\begin{aligned}
\text{where} \quad
I_2(z_1, z_2) &:= \langle g(z_1)g(z_2)\rangle.
\end{aligned}\label{70}
\end{equation}
These differential equations \eqref{5030} and \eqref{60} govern the macroscopic dynamics of learning. In addition, the generalization loss $\varepsilon_g$, the expectation of loss value $\varepsilon(\vect{\xi}) = \frac12 \|\vect{s} - \vect{t}\|^2$ over all input vectors $\vect{\xi}$, is represented as
\begin{equation}
\begin{aligned}
\varepsilon_g &= \langle \frac12 \|\vect{s}-\vect{t}\|^2 \rangle
= \frac12\left[ \sum_{p, q}^{M, M} F_{pq} I_2(y_p, y_q) + \sum_{p, q}^{K, K} D_{pq} I_2(x_p, x_q) \right.
\left. - 2\sum_{p, q}^{K, M} E_{pq} I_2(x_p, y_q) \right].
\end{aligned}\label{80}
\end{equation}

\subsection{Expectation terms}
Above we have determined the dynamics of order parameters as \eqref{5030}, \eqref{60} and \eqref{80}. However they have expectation terms $I_2(z_1, z_2)$, $I_3(z_1, z_2^{(e)}, z_3)$ and $I_4(z_1, z_2, z_3, z_4)$, where $z$s are either $x_i$ or $y_n$. By studying what distribution $\vect{z}$ follows, we can show that these expectation terms are dependent only on 1-st and $(e+1)$-th order parameters, namely, $Q^{(1)}, R^{(1)}, T^{(1)}$ and $Q^{(e+1)}, R^{(e+1)}, T^{(e+1)}$; for example,
\begin{align*}
I_3(x_i, x_j^{(e)}, y_p) &= \int dz_1dz_2dz_3 \  g'(z_1)z_2g(z_3) \ \mathcal{N}(\vect{z} | \vect{0}, \gyouretsu{Q_{ii}^{(1)}}{Q_{ij}^{(e+1)}}{R_{ip}^{(1)}}{Q_{ij}^{(e+1)}}{*}{R_{jp}^{(e+1)}}{R_{ip}^{(1)}}{R_{jp}^{(e+1)}}{T_{pp}^{(1)}})
\end{align*}
holds, where $*$ does not influence the value of this expression (see Supplementary Material A.1 for more detailed discussion). 
Thus, we see the `speed' of $e$-th order parameters (i.e. \eqref{5030} and \eqref{60}) only depends on 1-st and $(e+1)$-th order parameters, and the generalization error $\varepsilon_g$ (equation \eqref{80}) only depends on 1-st order parameters. Therefore, with denoting $(Q^{(e)}, R^{(e)}, T^{(e)})$ by $\Omega^{(e)}$ and $(D, E, F)$ by $\chi$, we can write
\begin{align*}
    \frac{d}{d\tilde{\alpha}}\Omega^{(e)} &= f^{(e)}(\Omega^{(1)}, \Omega^{(e+1)}, \chi), \qquad
    \frac{d}{d\tilde{\alpha}}\chi = g(\Omega^{(1)}, \chi), \quad\text{and}\quad
    \varepsilon_g = h(\Omega^{(1)}, \chi)
\end{align*}
with appropriate functions $f^{(e)}$, $g$ and $h$.
Additionally, a polynomial of $\Sigma$
\begin{align*}
    &P(\Sigma) := \prod_{i=1}^d (\Sigma - \lambda'_i I_N) = \sum_{e=0}^d c_e \Sigma^e \qquad
    \text{where}\quad \lambda'_1, \ldots, \lambda'_d \quad\text{are distinct eigenvalues of $\Sigma$}
\end{align*}
equals to 0, thus we get
\begin{equation}
\begin{aligned}
    \Omega^{(d)} = - \sum_{e=0}^{d-1} c_e \Omega^{(e)}.
\end{aligned}\label{10000}
\end{equation}
Using this relation, we can reduce $\Omega^{(d)}$ to expressions which contain only $\Omega^{(0)}, \ldots, \Omega^{(d-1)}$, therefore we can get a closed differential equation system with $\Omega^{(0)}, \ldots, \Omega^{(d-1)}$ and $\chi$.

In summary, our macroscopic system is closed with the following order parameters:
\begin{align*}
    &\text{Order variables}:
    \quad Q_{ij}^{(0)}, Q_{ij}^{(1)}, \ldots, Q_{ij}^{(d-1)},
    \quad R_{in}^{(0)}, R_{in}^{(1)}, \ldots, R_{in}^{(d-1)},
    \quad D_{ij}, E_{in} \\
    &\text{Order constants}:
    \quad T_{nm}^{(0)}, T_{nm}^{(1)}, \ldots, T_{nm}^{(d-1)},
    \quad F_{nm}.
    \qquad \text{($d$: number of distinct eigenvalues of $\Sigma$)}
\end{align*}
The order variables are governed by \eqref{5030} and \eqref{60}. For the lengthy full expressions of our macroscopic system for specific cases, see Supplementary Material A.2. 

\subsection{Dependency on input data covariance $\Sigma$}

\en{
The differential equation system we derived depends on $\Sigma$, through two ways; the coefficient $\mu_{e+1}$ of $O(\eta^2)$-term, and how ($d$)-th order parameters are expanded with lower order parameters (as \eqref{10000}). Specifically, the system only depends on the eigenvalue distribution of $\Sigma$.
}
\jp{
驚くべきことに，我々が導出したオーダパラメータの微分方程式系そのものは，
$O(\eta^2)$ の項の係数 $\mu_{e+1}$ を通してのみ $\Sigma$ に依存している．
すなわち，$\Sigma$ の固有値分布にのみ依存している．さらに，
学習係数 $\eta$ が小さい時には，我々の系は $\Sigma$ に全く依存しないものとなる．
ただし，実際には，$\Sigma$ は系の発展自体には影響を与えない代わりに，系の初期状態(i.e. initial values of the order parameters and values of order constants) に影響を与えていることに注意されたい．
}

\subsection{Evaluation of expectation terms for specific activation functions}
Expectation terms $I_2$, $I_3$ and $I_4$
can be analytically determined
for some activation functions $g$,
including sigmoid-like $g(x) = \textrm{erf}(x/\sqrt{2})$ (see \citet{saadsolla1995})
and $g(x) = \textrm{ReLU}(x)
$ (see \cite{yoshida2017statistical}).



\section{Analysis of numerical solutions of macroscopic differential equations}

\en{
In this section, we analyze numerically the order parameter system, derived in the previous section\footnote{
We executed all computations on a standard PC.}.
We assume that the second layers' weights of the student and the teacher, namely $w_i$ and $v_n$, are fixed to 1 (i.e. we consider the learning of soft-committee machine), and that $K$ and $M$ are equal to $2$, for simplicity. Here we think of sigmoid-like activation
$g(x) = \textrm{erf}(x/\sqrt{2})$. 
}\jp{
本節以降では，単純化のため，出力素子数 $O$ が 1 の場合を考える．また，生徒と教師のネットワークの第二層の重み $w_i$, $v_n$ は，全て 1 で固定されているものとする（すなわち，soft-committee machine の学習を考える）．活性化関数は sigmoid-like な $g(x) = \textrm{erf}(x/\sqrt{2})$ を考える．
}

\subsection{Consistency between macroscopic system and microscopic system}

\en{
First of all, we confirmed the consistency between the macroscopic system we derived and the original microscopic system. That is, we computed the dynamics of the generalization loss $\varepsilon_g$ in two ways: (i) by updating weights of the network with SGD \eqref{10} iteratively, and (ii) by solving numerically the differential equations \eqref{60} which govern the order parameters, and we confirmed that they accord with each other very well (\figref{fig_consistency}). Note that we set the initial values of order parameters in (ii) as values corresponding to initial weights used in (i). For dependence of the learning trajectory on the initial condition, see Supplementary Material A.3.
}\jp{
まず，前節で導出したマクロ系の妥当性を確かめるため，具体的な学習の発展をミクロとマクロの両方を用いて求め，それらが一致することを確かめた．すなわち，勾配法 \eqref{10} に従ってミクロな重みを更新することによって得られる汎化誤差 $\varepsilon_g$ のダイナミクスと，オーダパラメータの発展則 \eqref{5030}, \eqref{60} の微分方程式を解くことによって得られる汎化誤差  $\varepsilon_g$ のダイナミクスの一致を確認した（\figref{fig_consistency}）．ただし，マクロ系を解く際のオーダパラメータの初期値は，ミクロ系の実験に用いた初期重みから算出される値に設定していることに注意されたい．
}

\begin{figure}[htbp]
\begin{center}
\includegraphics[width=0.6\linewidth]{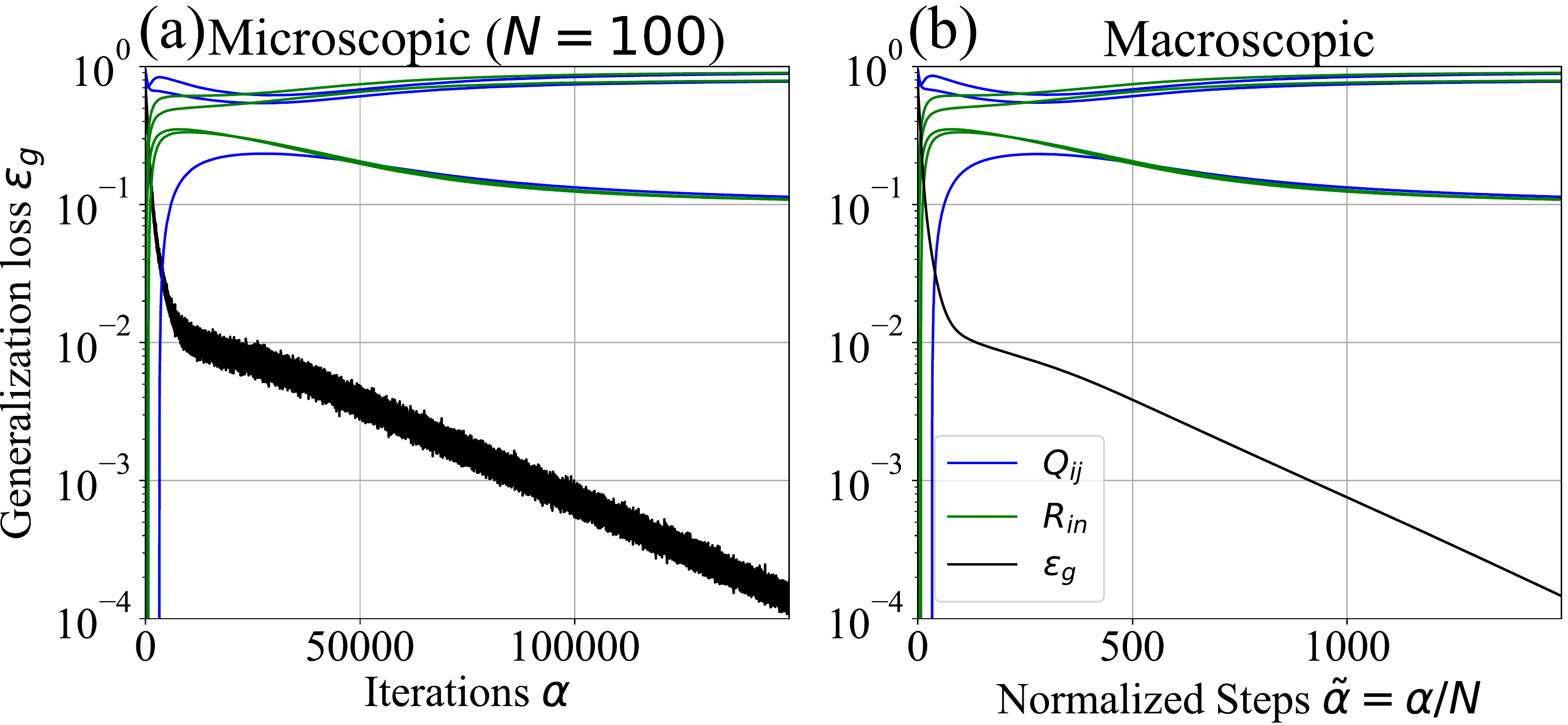}
\caption{
Example dynamics of generalization error $\varepsilon_g$ computed with (a) microscopic and (b) macroscopic system. Network size: $N$-$2$-$1$. Learning rate: $\eta=0.1$. Eigenvalues of $\Sigma$: $\lambda_1 = 0.4$ with multiplicity $0.5 N$, $\lambda_2 = 1.2$ with multiplicity $0.3 N$, and $\lambda_3 = 1.6$ with multiplicity $0.2 N$. Black lines: dynamics of $\varepsilon_g$. Blue lines: $Q_{11}, Q_{12}, Q_{22}$. Green lines: $R_{11}, R_{12}, R_{21}, R_{22}$.
}
\label{fig_consistency}
\end{center}
\end{figure}

\subsection{Case of scalar input covariance $\Sigma = \sigma I_N$}
\en{
As the simplest case, here we consider the case that the convariance matrix $\Sigma$ is proportional to unit matrix. In this case, $\Sigma$ has only one eigenvalue $\lambda = \mu_1$ of multiplicity $N$, then our order parameter system contains only parameters whose order is 0 ($e = 0$). For various values of $\mu_1$, we solved numerically the differential equations of order parameters \eqref{60} and plotted the time evolution of generalization loss $\varepsilon_g$ (\figref{fig_dynamics_with_various_mu1}(a)). From these plots, we quantified the lengths and heights of the plateaus as following: 
we regarded the system is plateauing if the decreasing speed of log-loss is smaller than half of its terminal converging speed, and we defined the height of the plateau as the median of loss values during plateauing. Quantified lengths and heights are plotted in \figref{fig_dynamics_with_various_mu1}(b)(c).
It indicates that the plateau length and height heavily depend on $\mu_1$, the input scale. Specifically, as $\mu_1$ decreases, the plateau rapidly becomes longer and lower.
Though smaller input data lead to longer plateaus, it also becomes lower and then inconspicuous. This tendency is consistent with \figref{fig_stwvid}(a)(b), since IRIS dataset has large $\mu_1$ ($\approx 15.9$) and MNIST has small $\mu_1$ ($\approx 0.112$). Considering this, the claim that the plateau phenomenon does not occur in learning of MNIST is controversy; this suggests the possibility that we are observing quite long and low plateaus. 

Note that \figref{fig_dynamics_with_various_mu1}(b) shows that the speed of growing of plateau length is larger than $O(1/\mu_1)$. This is contrast to the case of linear networks which have no activation; in that case, as $\mu_1$ decreases the speed of learning gets exactly $1/\mu_1$-times larger. In other words, this phenomenon is peculiar to nonlinear networks.
}
\jp{
最もシンプルなケースとして，入力の共分散行列が単位行列に比例する場合について考える．このとき，$\Sigma$ は 重複度 $N$ の 1 種類の固有値 $\lambda = \mu_1$ のみをもつ．様々な $\mu_1$ の値 に対して，オーダパラメータ微分方程式 \eqref{10} の数値解から求めた汎化誤差 $\varepsilon_g$ の時間発展を \figref{fig_dynamics_with_various_mu1} に示す．また \figref{fig_stats_with_various_mu1} には，\figref{fig_dynamics_with_various_mu1} に記した数値解から算出したプラトーの長さと高さを示す．
プラトーの長さが，$\mu_1$ に依存していることがわかる．具体的には，入力のノルム $\mu_1$ が小さくなるにつれ，プラトーは急激に長くかつ低くなる．ここで，\figref{fig_dynamics_with_various_mu1} (a) においてプラトーの長さが延長するレートは $O(1/\mu_1)$ よりも速いことに留意されたい．このことは，活性化関数のない線形ネットワークの場合において $\mu_1$ が小さくなるにつれ学習に要する時間が厳密に $O(1/\mu_1)$ 倍となることとは対照的である --- すなわち，非線形ネットワーク特有の現象といえる．
}

\begin{figure}[htbp]
\begin{center}
\includegraphics[width=0.8\linewidth]{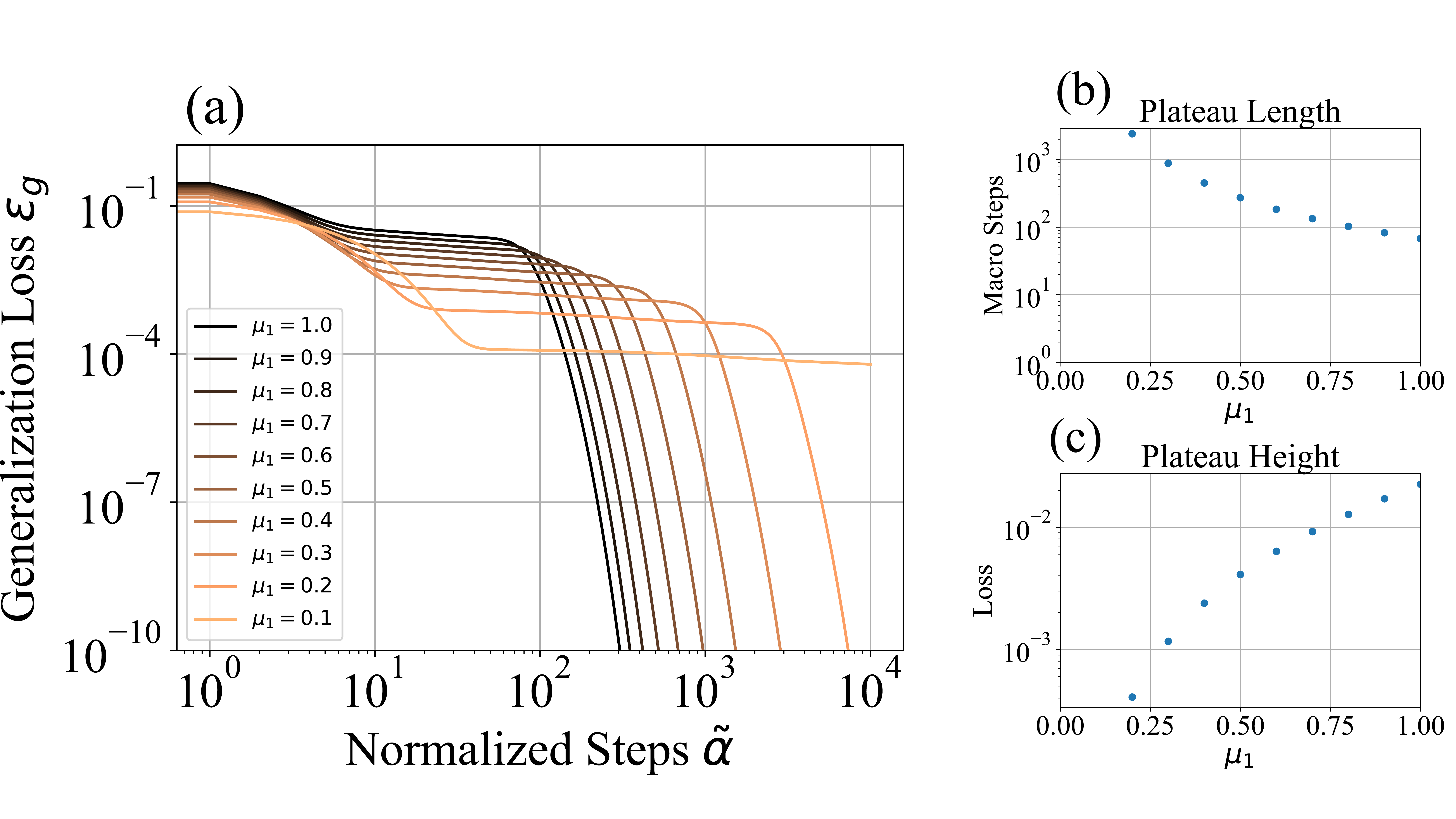}
\caption{
(a) Dynamics of generalization error $\varepsilon_g$ when input variance $\Sigma$ has only one eigenvalue $\lambda = \mu_1$ of multiplicity $N$. Plots with various values of $\mu_1$ are shown.
(b) Plateau length and (b) plateau height, quantified from (a).
}
\label{fig_dynamics_with_various_mu1}
\end{center}
\end{figure}


\subsection{Case of different input covariance $\Sigma$ with fixed $\mu_1$}
\en{
In the previous subsection we inspected the dependence of the learning dynamics on the first moment $\mu_1$ of the eigenvalues of the covariance matrix $\Sigma$. In this subsection, we explored the dependence of the dynamics on the higher moments of eigenvalues, under fixed first moment $\mu_1$.
}
\jp{
前小節では入力 $\vect{\xi}$ の共分散行列 $\Sigma$ の固有値分布の一次モーメント $\mu_1$ へ の学習ダイナミクスの依存性を調べたが，本小節では，一次モーメント $\mu_1$ が固定されている下でのより高次のモーメントへの学習ダイナミクスの依存性を調べる．
}

\en{
In this subsection, we consider the case in which the input covariance matrix $\Sigma$ has two distinct nonzero eigenvalues, $\lambda_1 = \mu_1 - \Delta\lambda/2$ and $\lambda_2 = \mu_1 + \Delta\lambda/2$, of the same multiplicity $N/2$ (\figref{fig_two_lambda}). With changing the control parameter $\Delta\lambda$, we can get eigenvalue distributions with various values of second moment $\mu_2 = \langle \lambda_i^2 \rangle$.
}\jp{
ここでも，非常にシンプルな設定として，入力の固有値分布が \figref{fig_two_lambda} に示すような分布をしたケースを考える．制御パラメータ $\Delta\lambda$ を変更することで，固有値分布の二次モーメント $\mu_2 = \langle \lambda_i^2 \rangle$ が変化する．
}

\begin{figure}[htbp]
\vspace{-0.4cm}
\begin{center}
\includegraphics[width=0.25\linewidth]{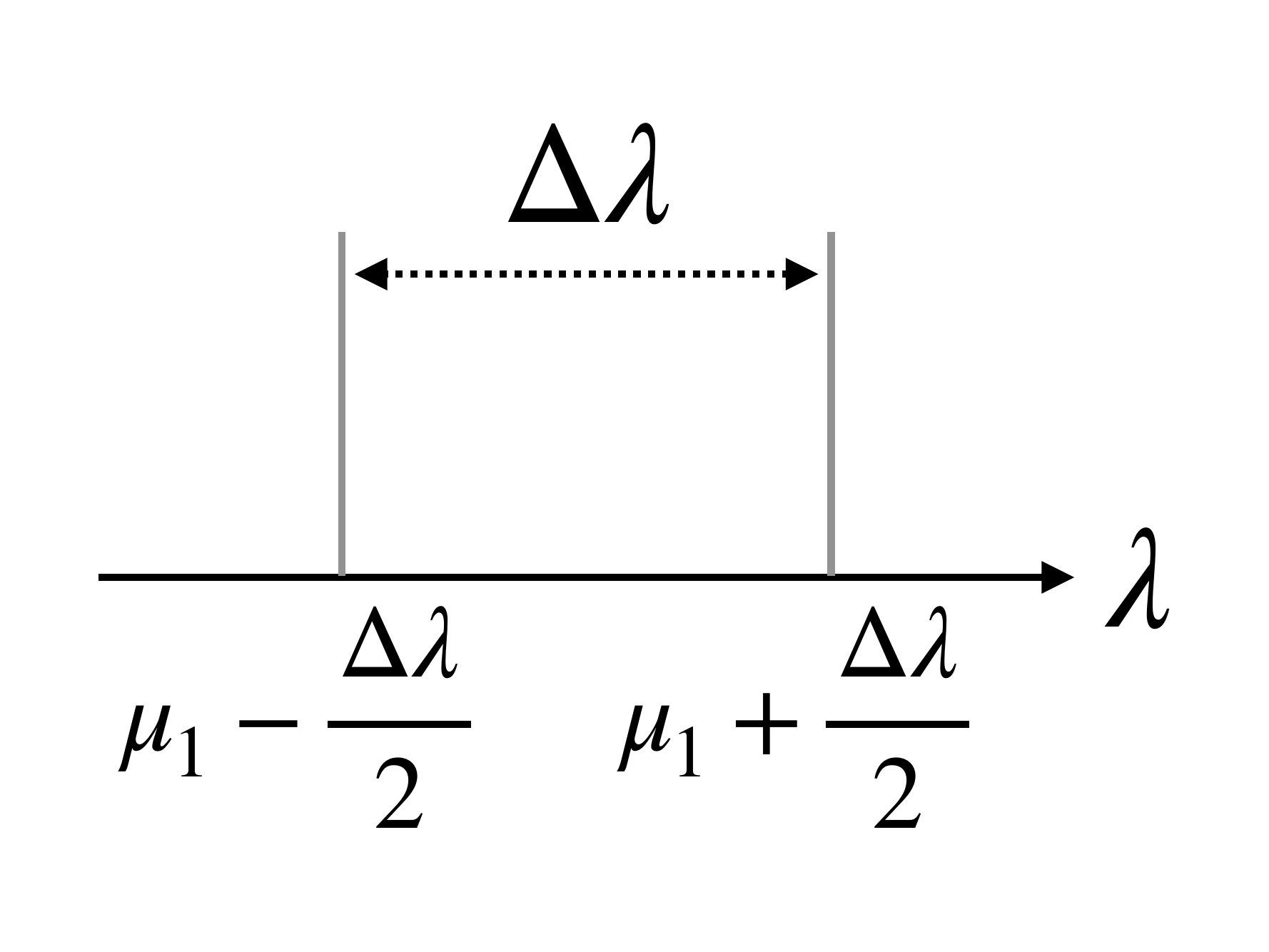}
\vspace{-0.4cm}
\caption{
Eigenvalue distribution with fixed $\mu_1$ parameterized by $\Delta \lambda$, which yields various $\mu_2$.
}
\label{fig_two_lambda}
\end{center}
\end{figure}

\en{
\figref{fig_dynamics_with_various_mu2}(a) shows learning curves with various $\mu_2$ while fixing $\mu_1$ to $1$. From these curves, we quantified the lengths and heights of the plateaus, and plotted them in \figref{fig_dynamics_with_various_mu2}(b)(c). These indicate that the length of the plateau shortens as $\mu_2$ becomes large. That is, the more the distribution of nonzero eigenvalues gets broaden, the more the plateau gets alleviated. 
}
\jp{
$\mu_1$ を固定しながら $\mu_2$ を様々に変化させた場合の，学習曲線を \figref{fig_dynamics_with_various_mu2} に示す．また \figref{fig_stats_with_various_mu2} には，\figref{fig_dynamics_with_various_mu2} に記した数値解から算出したプラトーの長さと高さを示す．これらの図から，$\mu_2$ が大きくなるにつれてプラトーの長さが短縮することがわかる．すなわち，この設定において，入力分布の分散共分散行列 $\Sigma$ の非ゼロの固有値たちのなす分布が大きな分散をもつほど，プラトーは軽減される傾向にあることがわかる．
}

\begin{figure}[htbp]
\begin{center}
\includegraphics[width=0.8\linewidth]{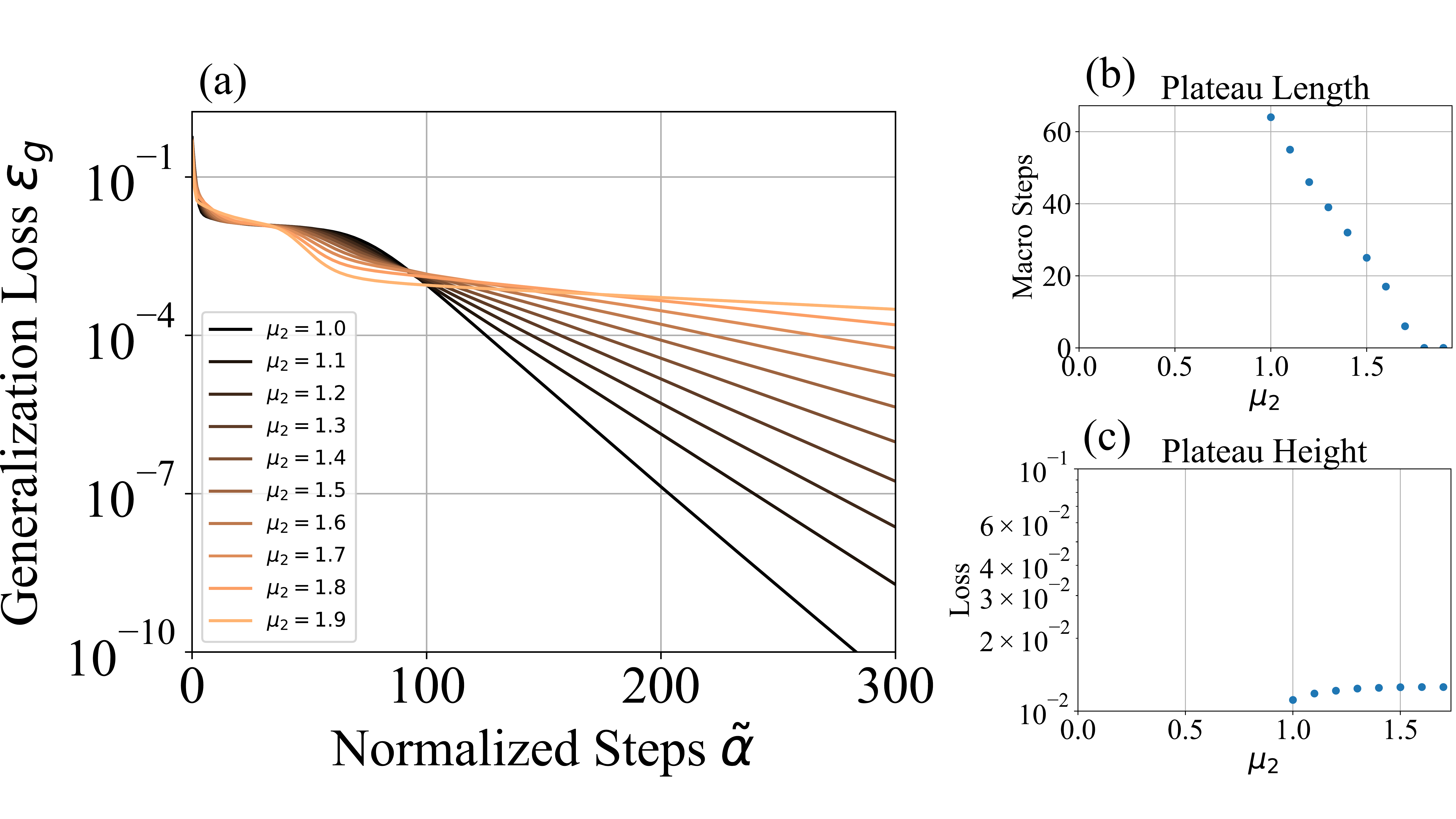}
\caption{
(a) Dynamics of generalization error $\varepsilon_g$ when input variance $\Sigma$ has two eigenvalues $\lambda_{1,2} = \mu_1 \pm \Delta\lambda/2$ of multiplicity $N/2$. Plots with various values of $\mu_2$ are shown.
(b) Plateau length and (c) plateau height, quantified from (a).
}
\label{fig_dynamics_with_various_mu2}
\end{center}
\end{figure}


\section{Conclusion}
Under the statistical mechanical formulation of learning in the two-layered perceptron, we showed that macroscopic equations can be derived even when the statistical properties of the input are generalized. We showed that the dynamics of learning depends only on the eigenvalue distribution of the covariance matrix of the input data.
By numerically analyzing the macroscopic system, it is shown that the statistics of input data dramatically affect the plateau phenomenon.

Through this work, we explored the gap between theory and reality; though the plateau phenomenon is theoretically predicted to occur by the general symmetrical structure of neural networks, it is seldom observed in practice. However, more extensive researches are needed to fully understand the theory underlying the plateau phenomenon in practical cases.

\section*{Acknowledgement}
This work was supported by JSPS KAKENHI Grant-in-Aid for Scientific
Research(A) (No. 18H04106).

\bibliographystyle{plainnat}
\bibliography{reflist}

}\supplementary{

\title{\titlestr \\ (Supplementary Material)}

\maketitle

\setcounter{section}{0}
\renewcommand{\thesection}{A.\arabic{section}}
\renewcommand{\thesubsection}{A.\arabic{section}.\arabic{subsection}}
\renewcommand{\thetable}{A.\arabic{table}}
\renewcommand{\thefigure}{A.\arabic{figure}}

\section{Properties of expectation term $I_2$, $I_3$ and $I_4$}

The differential equations of learning dynamics (3) and (5) in the main text have expectation terms, $I_2(z_1, z_2), I_3(z_1, z_2, z_3)$
and $I_4(z_1, z_2, z_3, z_4)$. Since their $z$s are either $x^{(e)}_i = \vect{\xi}^T \Sigma^e \vect{J}_i$ or $y^{(e)}_n = \vect{\xi}^T \Sigma^e \vect{B}_n$, any tuple $(z_1, z_2, \ldots)$ follows multivatiate normal distribution $\mathcal{N}(\vect{z} | 0, \langle \vect{z}\cdot\vect{z}^T \rangle)$ when $N \to \infty$ by generalized central limit theorem, provided that the input $\vect{\xi}$ has zero mean and finite covariance. Thus the expectation terms only depend on the covariance matrix $\langle \vect{z}\cdot\vect{z}^T \rangle$, and their elements can be calculated as
$\langle x^{(e)}_i x^{(f)}_j \rangle = Q_{ij}^{(e+f+1)}$, $\langle x^{(e)}_i y^{(f)}_n \rangle = R_{in}^{(e+f+1)}$ and $\langle y^{(e)}_n y^{(f)}_m \rangle = T_{nm}^{(e+f+1)}$. For example,
\begin{align*}
I_2(x_i, y_p) &= \int dz_1dz_2 \  g(z_1)g(z_2) \ \mathcal{N}(\vect{z} | \vect{0}, \gyouretuthree{Q_{ii}^{(1)}}{R_{ip}^{(1)}}{T_{pp}^{(1)}}), \\
I_3(x_i, x_j^{(e)}, y_p) &= \int dz_1dz_2dz_3 \  g'(z_1)z_2g(z_3) \ \mathcal{N}(\vect{z} | \vect{0}, \gyouretsu{Q_{ii}^{(1)}}{Q_{ij}^{(e+1)}}{R_{ip}^{(1)}}{}{Q_{jj}^{(2e+1)}}{R_{jp}^{(e+1)}}{}{}{T_{pp}^{(1)}}), \\
I_4(x_i, x_j, y_p, y_q) &= \int dz_1dz_2dz_3dz_4 \  g(z_1)g(z_2)g(z_3)g(z_4) \ \mathcal{N}(\vect{z} | \vect{0}, \gyouretuseveno{R\o_{ip}}{R\o_{iq}}{R\o_{jp}}{R\o_{jq}}{T\o_{pp}}{T\o_{pq}}{T\o_{qq}}).
\end{align*}
Note that all the covariance matrix is symmetric. Their left-bottom sides are not shown for notational simplicity.
Substituting these for $I$s shown in equations (3) and (5) in the main text, we see that the `speed' of $e$-th order parameters can be dependent only on $1$-st, $(e+1)$-th, and $(2e+1)$-th order parameters. 

Here we prove the following proposition, in order to show that the `speed' of $e$-th order parameters are not dependent on $(2e+1)$-th order parameters. 

\noindent\textbf{Proposition.}\quad
The expectation term $I_3(z_1, z_2, z_3) := \int  dz_1dz_2dz_3 \ g'(z_1)z_2g(z_3) \ \mathcal{N}(\vect{z} | \vect{0}, C)$ does not depend on $C_{22}$.    

\en{
\noindent\textbf{Proof.}\quad Since $C$ is positive-semidefinite, we can write $C = VV^T$ for some squared matrix $V$. Thus, when $\vect{\xi} \sim \mathcal{N}(0, I_N)$, $A\vect{\xi} \sim \mathcal{N}(0, C)$ holds. Therefore, we can regard that $z_i (i = 1, 2, 3)$ is generated by $z_i = \vect{v}_i^T \vect{\xi}$ where $\vect{v}_i$ is $i$-th row vector of $V$ and $\vect{\xi}$ follows the standard normal distribution. 

We can write $\vect{v}_2 = c_1 \vect{v}_1 + c_3 \vect{v}_3 + \vect{v}^\perp$ for some coefficient $c_1, c_3 \in \mathbb{R}$ and some vector $\vect{v}^\perp$ perpendicular to $\vect{v}_1$ and $\vect{v}_3$. Then $I_3$ is written as 
\begin{align*}
I_3(z_1, z_2, z_3)
= \langle g'(z_1) z_2 g(z_3) \rangle
=  c_1 \langle g'(z_1) z_1 g(z_3) \rangle + c_3 \langle g'(z_1) z_3 g(z_3) \rangle + \langle g'(z_1) \vect{v}^{\perp T}\vect{\xi} g(z_3) \rangle.
\end{align*}
Since $\vect{\xi} \sim \mathcal{N}(0, I_N)$ and $\vect{v}^\perp \perp \vect{v}_1, \vect{v}_3$ hold, $(z_1, z_3)$ and $\vect{v}^{\perp T}\vect{\xi}$ is independent. Therefore the third term in the right hand side of the equation above is
\begin{align*}
\langle g'(z_1) \vect{v}^{\perp T}\vect{\xi} g(z_3) \rangle = \langle g'(z_1) g(z_3) \rangle \langle \vect{v}^{\perp T}\vect{\xi} \rangle = 0.
\end{align*}
In addition, we can determine $c_1$ and $c_3$ by solving
\begin{align*}
C_{12}
&= \vect{v}_2^T\vect{v}_1 = (c_1 \vect{v}_1^T+ c_3 \vect{v}_3^T + \vect{v}^{\perp T})\vect{v}_1 
= c_1 C_{11} + c_3 C_{13} \quad\text{and} \\
C_{23}
&= \vect{v}_2^T\vect{v}_3 = (c_1 \vect{v}_1^T+ c_3 \vect{v}_3^T + \vect{v}^{\perp T})\vect{v}_3
= c_1 C_{13} + c_3 C_{33}.
\end{align*}
Together with these, we get
\begin{align*}
I_3(z_1, z_2, z_3) = \frac{(C_{12}C_{33}-C_{13}C_{23}) \ I_3(z_1, z_1, z_3)
+ (C_{11}C_{23}-C_{12}C_{13}) \ I_3(z_1, z_3, z_3)}{C_{11}C_{33}-C_{13}^2},
\end{align*}
which shows that $I_3$ is independent to $C_{22}$.  $\blacksquare$
}
\jp{
ある $n$ 次元定ベクトル $\vect{z}_i$ が存在し、$\vect{z}_i\cdot\vect{\xi} = x_i$ と書くことができる
（ただし $\vect{\xi}$ は $n$ 次元標準正規分布に従う）\footnote{多分 $C$ か $C^{-1}$の固有ベクトル}。
このとき、$I_3$ は次のように簡約化できる：$\vect{z}_2 = c_1 \vect{z}_1 + c_3 \vect{z}_3 + \vect{z}^\perp$ と書くと
（ただし $\vect{z}^\perp$ は $\vect{z}_1$, $\vect{z}_3$ と直交）、
\begin{align*}
\langle g'(x_1) x_2 g(x_3) \rangle
=  c_1 \langle g'(x_1) x_1 g(x_3) \rangle + c_3 \langle g'(x_1) x_3 g(x_3) \rangle + \langle g'(x_1) \vect{z}^\perp\cdot\vect{\xi} g(x_3) \rangle
\end{align*}
ここで第三項は 0 である。なぜなら $(x_1, x_3)$ と $x_2$ は無相関であり、それゆえ（多変量正規分布に従うことから）独立であるため
$\langle g'(x_1) \vect{z}^\perp\cdot\vect{\xi} g(x_3) \rangle = \langle g'(x_1) g(x_3) \rangle \langle \vect{z}^\perp\cdot\vect{\xi} \rangle = 0$
である。
また、$\vect{z}_2 = c_1 \vect{z}_1 + c_3 \vect{z}_3 + \vect{z}^\perp$ に $\vect{z}_1$ と $\vect{z}_3$ を掛けて得られる連立方程式を
解くことにより $(c_1, c_3)$ が求められる。結局
\begin{align*}
I_3(x_1, x_2, x_3) = \frac{(C_{12}C_{33}-C_{13}C_{23}) \ I_3(x_1, x_1, x_3)
+ (C_{11}C_{23}-C_{12}C_{13}) \ I_3(x_1, x_3, x_3)}{C_{11}C_{33}-C_{13}^2}
\end{align*}
が得られる。
}

\section{Full expression of order parameter system}

Here we describe the whole system of the order parameters, with specific eigenvalue distribution of $\Sigma$.


\subsection{Case with $\Sigma = \sigma I_N$}

In this case, the order parameters are
\begin{align*}
    &\text{Order variables}: 
    \qquad Q_{ij}^{(0)}, 
    \qquad R_{in}^{(0)}, 
    \qquad D_{ij}, E_{in} \\
    &\text{Order constants}:
    \qquad T_{nm}^{(0)},
    \qquad F_{nm}.
\end{align*}
Note that $Q_{ij}^{(1)}$ is identical to $Q_{ij}^{(0)}$. This is same for $R$ and $T$.
The order parameter system is described as following, with omitting ${}^{(0)}$-s for notational simplicity:
\begin{equation}
\begin{aligned}
\frac{dQ_{ij}}{d\tilde{\alpha}}
&= \eta \left[ \sum_{p=1}^M E_{ip} I_3\gyouretusix{Q_{ii}}{Q_{ij}}{R_{ip}}{Q_{jj}}{R_{jp}}{T_{pp}} - \sum_{p=1}^K D_{ip} I_3\gyouretusix{Q_{ii}}{Q_{ij}}{Q_{ip}}{Q_{jj}}{Q_{jp}}{Q_{pp}} \right. \\
&+ \left. \sum_{p=1}^M E_{jp} I_3\gyouretusix{Q_{jj}}{Q_{ji}}{R_{jp}}{Q_{ii}}{R_{ip}}{T_{pp}} - \sum_{p=1}^K D_{jp} I_3\gyouretusix{Q_{jj}}{Q_{ji}}{Q_{jp}}{Q_{ii}}{Q_{ip}}{Q_{pp}} \right] \\
&+ \eta^2 \left[
\sum_{p, q}^{K, K} D_{ip}D_{jq} I_4\gyouretuseven{Q_{ip}}{Q_{iq}}{Q_{jp}}{Q_{jq}}{Q_{pp}}{Q_{pq}}{Q_{qq}}
+ \sum_{p, q}^{M, M} E_{ip}E_{jq} I_4\gyouretuseven{R_{ip}}{R_{iq}}{R_{jp}}{R_{jq}}{T_{pp}}{T_{pq}}{T_{qq}} \right. \\
&\qquad \left. - \sum_{p, q}^{K, M} D_{ip}E_{jq} I_4\gyouretuseven{Q_{ip}}{R_{iq}}{Q_{jp}}{R_{jq}}{Q_{pp}}{R_{pq}}{T_{qq}}
- \sum_{p, q}^{M, K} E_{ip}D_{jq} I_4\gyouretuseven{R_{ip}}{Q_{iq}}{R_{jp}}{Q_{jq}}{T_{pp}}{R_{pq}}{Q_{qq}} \right] , \\
\frac{dR_{in}}{d\tilde{\alpha}}
&= \eta \left[ \sum_{p=1}^M E_{ip} I_3\gyouretusix{Q_{ii}}{R_{in}}{R_{ip}}{T_{nn}}{T_{np}}{T_{pp}} - \sum_{p=1}^K D_{ip} I_3\gyouretusix{Q_{ii}}{R_{in}}{Q_{ip}}{T_{nn}}{R_{pn}}{Q_{pp}} \right]
\end{aligned}\label{30}
\end{equation}
and 
\begin{equation}
\begin{aligned}
\frac{dD_{ij}}{d\tilde{\alpha}}
&= \eta \left[ \sum_{p=1}^M E_{ip} I_2\gyouretuthree{Q_{jj}}{R_{jp}}{T_{pp}} - \sum_{p=1}^K D_{ip} I_2\gyouretuthree{Q_{jj}}{Q_{jp}}{Q_{pp}} \right. \\
&\qquad \left. +  \sum_{p=1}^M E_{jp} I_2\gyouretuthree{Q_{ii}}{R_{ip}}{T_{pp}} - \sum_{p=1}^K D_{jp} I_2\gyouretuthree{Q_{ii}}{Q_{ip}}{Q_{pp}} \right], \\
\frac{dE_{in}}{d\tilde{\alpha}}
&= \eta \left[ \sum_{p=1}^M F_{pn} I_2\gyouretuthree{Q_{ii}}{R_{ip}}{T_{pp}} - \sum_{p=1}^K E_{pn} I_2\gyouretuthree{Q_{jj}}{R_{jp}}{T_{pp}} \right]
\end{aligned}\label{8060}
\end{equation},
\begin{equation}
\begin{aligned}
\text{where} \quad
I_2(C) &= \frac{2}{\pi}\arcsin\frac{C_{12}}{\sqrt{1+C_{11}}\sqrt{1+C_{22}}}, \\
I_3(C) &= \frac{2}{\pi}\cdot\frac{1}{\sqrt{(1+C_{11})(1+C_{33})-C_{13}^2}}\frac{C_{23}(1+C_{11})-C_{12}C_{13}}{1+C_{11}}, \\
I_4(C) &= \frac{4}{\pi^2}\cdot\frac{1}{\sqrt{1+2C_{11}}}
\arcsin\frac{(1+2C_{11})C_{23}-2C_{12}C_{13}}{
\sqrt{(1+2C_{11})(1+C_{22}) - 2 C_{12}^2}
\sqrt{(1+2C_{11})(1+C_{33}) - 2 C_{13}^2}
}
\end{aligned}\label{8040}
\end{equation}
for $g(x) = \textrm{erf}(x/\sqrt{2})$ activation, as \citet{saadsolla1995} showed.

\subsection{Case with $\Sigma$ which has two distinct eigenvalues, $\lambda_1$ of multiplicity $r_1N$ and $\lambda_2$ of multiplicity $r_2N$}

In this case, the order parameters are
\begin{align*}
    &\text{Order variables}:
    \qquad Q_{ij}^{(0)}, Q_{ij}^{(1)},
    \qquad R_{in}^{(0)}, R_{in}^{(1)},
    \qquad D_{ij}, E_{in} \\
    &\text{Order constants}:
    \qquad T_{nm}^{(0)}, T_{nm}^{(1)},
    \qquad F_{nm}.
\end{align*}
Since $\Sigma^2 - (\lambda_1 + \lambda_2)\Sigma + \lambda_1\lambda_2 I_N = 0$,
the relation $Q\w_{ij} = (\lambda_1 + \lambda_2)Q\o_{ij} - \lambda_1\lambda_2 Q\z_{ij}$ holds. This is same for $R$ and $T$.
Then the order parameter system is described as following:
\begin{equation}
\begin{aligned}
\frac{dQ\z_{ij}}{d\tilde{\alpha}}
&= \eta \left[ \sum_{p=1}^M E_{ip} I_3\gyouretusix{Q\o_{ii}}{Q\o_{ij}}{R\o_{ip}}{Q\o_{jj}}{R\o_{jp}}{T\o_{pp}} - \sum_{p=1}^K D_{ip} I_3\gyouretusix{Q\o_{ii}}{Q\o_{ij}}{Q\o_{ip}}{Q\o_{jj}}{Q\o_{jp}}{Q\o_{pp}} \right. \\
&+ \left. \sum_{p=1}^M E_{jp} I_3\gyouretusix{Q\o_{jj}}{Q\o_{ji}}{R\o_{jp}}{Q\o_{ii}}{R\o_{ip}}{T\o_{pp}} - \sum_{p=1}^K D_{jp} I_3\gyouretusix{Q\o_{jj}}{Q\o_{ji}}{Q\o_{jp}}{Q\o_{ii}}{Q\o_{ip}}{Q\o_{pp}} \right] \\
&+ \eta^2 (r_1\lambda_1 + r_2\lambda_2) \left[
\sum_{p, q}^{K, K} D_{ip}D_{jq} I_4\gyouretuseveno{Q\o_{ip}}{Q\o_{iq}}{Q\o_{jp}}{Q\o_{jq}}{Q\o_{pp}}{Q\o_{pq}}{Q\o_{qq}}
+ \sum_{p, q}^{M, M} E_{ip}E_{jq} I_4\gyouretuseveno{R\o_{ip}}{R\o_{iq}}{R\o_{jp}}{R\o_{jq}}{T\o_{pp}}{T\o_{pq}}{T\o_{qq}} \right. \\
&\qquad \left. - \sum_{p, q}^{K, M} D_{ip}E_{jq} I_4\gyouretuseveno{Q\o_{ip}}{R\o_{iq}}{Q\o_{jp}}{R\o_{jq}}{Q\o_{pp}}{R\o_{pq}}{T\o_{qq}}
- \sum_{p, q}^{M, K} E_{ip}D_{jq} I_4\gyouretuseveno{R\o_{ip}}{Q\o_{iq}}{R\o_{jp}}{Q\o_{jq}}{T\o_{pp}}{R\o_{pq}}{Q\o_{qq}} \right] , \\
\frac{dQ\o_{ij}}{d\tilde{\alpha}}
&= \eta \left[ \sum_{p=1}^M E_{ip} I_3\gyouretusix{Q\o_{ii}}{
    (\lambda_1 + \lambda_2)Q\o_{ij} - \lambda_1\lambda_2 Q\z_{ij}
}{R\o_{ip}}{Q\h_{jj}}{
    (\lambda_1 + \lambda_2)R\o_{jp} - \lambda_1\lambda_2 R\z_{jp}
}{T\o_{pp}} \right. \\
&- \left. \sum_{p=1}^K D_{ip} I_3\gyouretusix{Q\o_{ii}}{
    (\lambda_1 + \lambda_2)Q\o_{ij} - \lambda_1\lambda_2 Q\z_{ij}
}{Q\o_{ip}}{Q\h_{jj}}{
    (\lambda_1 + \lambda_2)Q\o_{jp} - \lambda_1\lambda_2 Q\z_{jp}
}{Q\o_{pp}} \right. \\
&+ \left. \sum_{p=1}^M E_{jp} I_3\gyouretusix{Q\o_{jj}}{
    (\lambda_1 + \lambda_2)Q\o_{ji} - \lambda_1\lambda_2 Q\z_{ji}
}{R\o_{jp}}{Q\h_{ii}}{
    (\lambda_1 + \lambda_2)R\o_{ip} - \lambda_1\lambda_2 R\z_{ip}
}{T\o_{pp}} \right. \\
&- \left. \sum_{p=1}^K D_{jp} I_3\gyouretusix{Q\o_{jj}}{
    (\lambda_1 + \lambda_2)Q\o_{ji} - \lambda_1\lambda_2 Q\z_{ji}
}{Q\o_{jp}}{Q\h_{ii}}{
    (\lambda_1 + \lambda_2)Q\o_{ip} - \lambda_1\lambda_2 Q\z_{ip}
}{Q\o_{pp}} \right] \\
&+ \eta^2 (r_1\lambda_1^2 + r_2\lambda_2^2) \left[
\sum_{p, q}^{K, K} D_{ip}D_{jq} I_4\gyouretuseveno{Q\o_{ip}}{Q\o_{iq}}{Q\o_{jp}}{Q\o_{jq}}{Q\o_{pp}}{Q\o_{pq}}{Q\o_{qq}}
+ \sum_{p, q}^{M, M} E_{ip}E_{jq} I_4\gyouretuseveno{R\o_{ip}}{R\o_{iq}}{R\o_{jp}}{R\o_{jq}}{T\o_{pp}}{T\o_{pq}}{T\o_{qq}} \right. \\
&\qquad \left. - \sum_{p, q}^{K, M} D_{ip}E_{jq} I_4\gyouretuseveno{Q\o_{ip}}{R\o_{iq}}{Q\o_{jp}}{R\o_{jq}}{Q\o_{pp}}{R\o_{pq}}{T\o_{qq}}
- \sum_{p, q}^{M, K} E_{ip}D_{jq} I_4\gyouretuseveno{R\o_{ip}}{Q\o_{iq}}{R\o_{jp}}{Q\o_{jq}}{T\o_{pp}}{R\o_{pq}}{Q\o_{qq}} \right] , \\
\frac{dR\z_{in}}{d\tilde{\alpha}}
&= \eta \left[ \sum_{p=1}^M E_{ip} I_3\gyouretusix{Q\o_{ii}}{R\o_{in}}{R\o_{ip}}{T\o_{nn}}{T\o_{np}}{T\o_{pp}} - \sum_{p=1}^K D_{ip} I_3\gyouretusix{Q\o_{ii}}{R\o_{in}}{Q\o_{ip}}{T\o_{nn}}{R\o_{pn}}{Q\o_{pp}} \right], \\
\frac{dR\o_{in}}{d\tilde{\alpha}}
&= \eta \left[ \sum_{p=1}^M E_{ip} I_3\gyouretusix{Q\o_{ii}}{(\lambda_1 + \lambda_2)R\o_{in} - \lambda_1\lambda_2R\z_{in}}{R\o_{ip}}{T\o_{nn}}{(\lambda_1 + \lambda_2)T\o_{np} - \lambda_1\lambda_2T\z_{np}}{T\o_{pp}} \right. \\
&- \left. \sum_{p=1}^K D_{ip} I_3\gyouretusix{Q\o_{ii}}{(\lambda_1 + \lambda_2)R\o_{in} - \lambda_1\lambda_2R\z_{in}}{Q\o_{ip}}{T\o_{nn}}{(\lambda_1 + \lambda_2)R\o_{pn} - \lambda_1\lambda_2R\z_{pn}}{Q\o_{pp}} \right]
\end{aligned}\label{9030}
\end{equation},
and
\begin{equation}
\begin{aligned}
\frac{dD_{ij}}{d\tilde{\alpha}}
&= \eta \left[ \sum_{p=1}^M E_{ip} I_2\gyouretuthree{Q\o_{jj}}{R\o_{jp}}{T\o_{pp}} - \sum_{p=1}^K D_{ip} I_2\gyouretuthree{Q\o_{jj}}{Q\o_{jp}}{Q\o_{pp}} \right. \\
&\qquad \left. +  \sum_{p=1}^M E_{jp} I_2\gyouretuthree{Q\o_{ii}}{R\o_{ip}}{T\o_{pp}} - \sum_{p=1}^K D_{jp} I_2\gyouretuthree{Q\o_{ii}}{Q\o_{ip}}{Q\o_{pp}} \right], \\
\frac{dE_{in}}{d\tilde{\alpha}}
&= \eta \left[ \sum_{p=1}^M F_{pn} I_2\gyouretuthree{Q\o_{ii}}{R\o_{ip}}{T\o_{pp}} - \sum_{p=1}^K E_{pn} I_2\gyouretuthree{Q\o_{jj}}{R\o_{jp}}{T\o_{pp}} \right]
\end{aligned}\label{9060}
\end{equation}.

\section{Dependence of learning trajectory on initial conditions on macroscopic parameters}

\begin{figure}[htbp]
\begin{center}
\vspace{-5mm}
\includegraphics[width=0.8\linewidth]{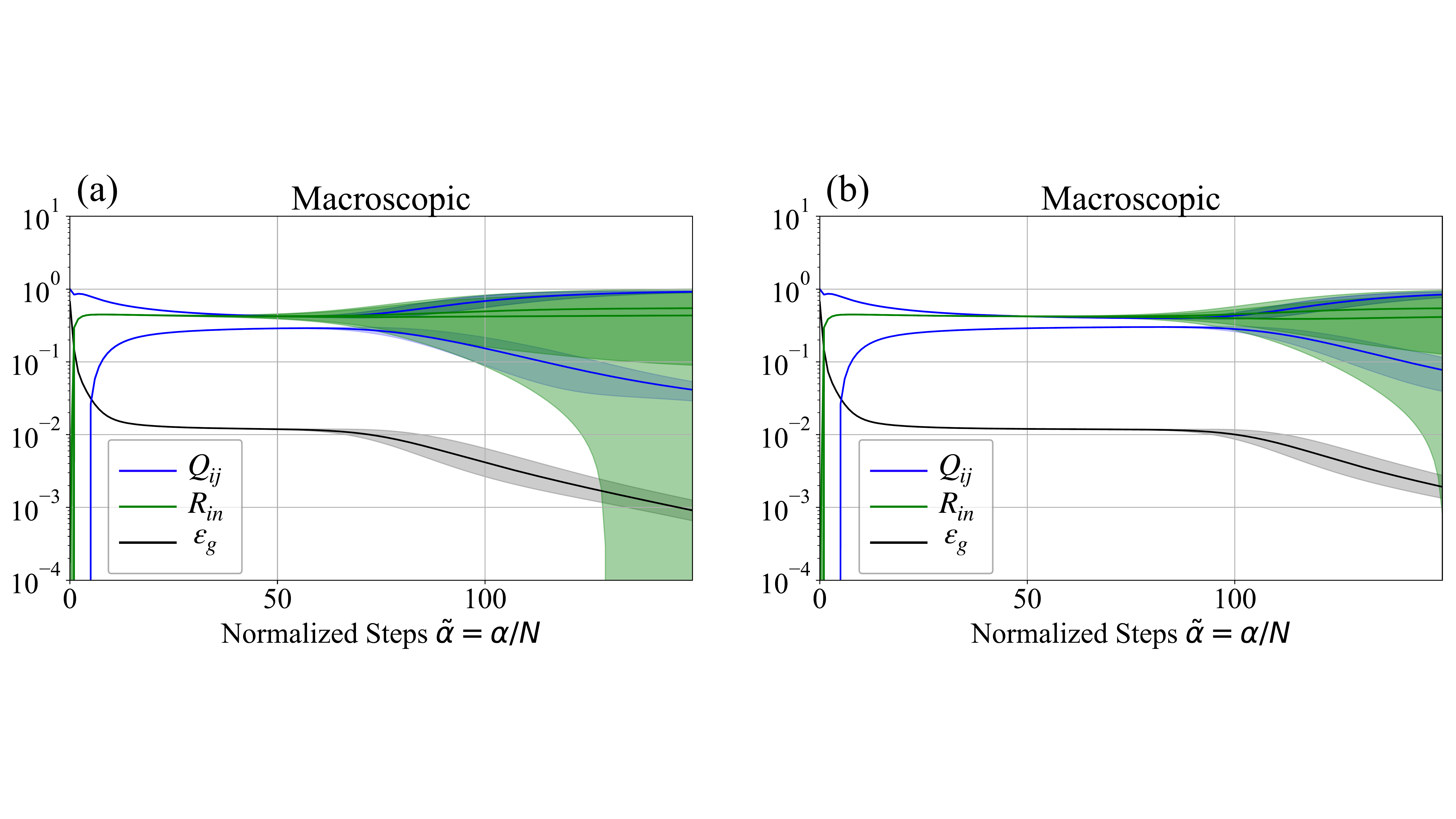}
\vspace{-10mm}
\caption{
Dynamics of generalization error $\varepsilon_g$ and order parameters $Q_{ij}$ and $R_{in}$ computed with macroscopic system, and its variability by random weight initialization. Network size: $N$-$2$-$1$. Learning rate: $\eta=0.1$. Eigenvalues of $\Sigma$: $\lambda_1 = 0.3$ with multiplicity $0.5 N$, $\lambda_2 = 1.7$ with multiplicity $0.5 N$. Black lines: dynamics of $\varepsilon_g$. Blue lines: $Q_{11}, Q_{12}, Q_{22}$. Green lines: $R_{11}, R_{12}, R_{21}, R_{22}$. (a) $N = 10^5$, (b) $N = 10^7$. In both figures, solid curves and shades represent mean and standard deviation of 100 trials, respectively (note that mean and standard deviation of loss are computed in logarithmic scale).
}
\label{fig_macro1_cloud}
\end{center}
\end{figure}

In the statistical mechanical formulation, by considering $N$ as large, the dynamics of the system is reduced to macroscopic differential equations with small ($N$-independent) dimensions. The macroscopic system we derived is deterministic in the sense that randomness brought by stochastic gradient descent is vanished.
However, note that the trajectory of the macroscopic state can vary in accordance with its initial condition. \figref{fig_macro1_cloud} shows this variability with shades.

How does the initial condition affect the learning trajectory? Consider a typical initialization that the microscopic parameters $\vect{J}_1$, $\vect{J}_2$, $\vect{B}_1$ and $\vect{B}_2$ are initialized as $(\vect{J}_i)_k, (\vect{B}_n)_k \stackrel{\textrm{i.i.d.}}{\sim} \mathcal{N}(0, 1/N)$. Then the mean and variance of corresponding initial macroscopic parameters $Q$, $R$ and $T$ are 
\begin{align*}
    \mathbb{E}[Q_{ii}^{(e)}] &= \mu_e, \quad
    \mathbb{V}[Q_{ii}^{(e)}] = \frac{3 \mu_{2e}}{N}, \quad
    \mathbb{E}[Q_{ij}^{(e)}] = 0, \quad
    \mathbb{V}[Q_{ij}^{(e)}] = \frac{\mu_{2e}}{N}, \\
    \mathbb{E}[R_{in}^{(e)}] &= 0, \quad
    \mathbb{V}[R_{in}^{(e)}] = \frac{\mu_{2e}}{N}, \\
    \mathbb{E}[T_{nn}^{(e)}] &= \mu_e, \quad
    \mathbb{V}[T_{nn}^{(e)}] = \frac{3 \mu_{2e}}{N}, \quad
    \mathbb{E}[T_{nm}^{(e)}] = 0, \quad
    \mathbb{V}[T_{nm}^{(e)}] = \frac{\mu_{2e}}{N}
\end{align*}
With $N \to \infty$, these probabilistic parameters converge to $(Q^{(e)}, R^{(e)}, T^{(e)}) = (\mu_e I_K, 0, \mu_e I_M)$. However, the solution trajectory starting from just $(\mu_e I_K, 0, \mu_e I_M)$ cannot break the weight symmetry at all. To argue practical learning trajectory, we have to consider the initial value slightly off from that point. How close the initial condition is to that point affects how long it takes to break the weight symmetry, that is, the plateau length. This is why \figref{fig_macro1_cloud} (b) with $N = 10^7$ exhibits plateau slightly longer than that of \figref{fig_macro1_cloud} (a) with $N = 10^5$.

\en{
}\jp{
・この節は本当に必要か？次の仕事にしても良いくらいのボリュームがあるのでは？

本節では，MNIST においてプラトーが本当に生じているかどうかを考えるにあたって，一つの観点を与える．

プラトー現象の主要因として，loss landscape 上の singularity がある．よく知られている singularity として，第一層の重みが degenerated していたり消失していたりする singularity がある (Fukumizu & Amari)．これらの時には，第二層の重みに不定性が生じ，これが loss landscape に特異な構造をもたらす．

ここで自然な問いが生じる．MNIST dataset の学習中に，このような中間素子の degeneration が生じているだろうか．

MNIST の学習中の weight overlap や weight vanishing が生じているかを調べるため，$Q = JJ$ のノルムやオーバーラップを調べた．

しかしながら，weight overlap や weight vanishing は中間素子の degeneration が生じる十分条件であり，必要条件ではない．任意の入力データ $x$ に対して $g(J_1x) = g(J_2x)$ となるならば，$J_1 = J_2$ でなくとも，実質的に素子 1 と素子 2 は縮退している．MNIST をはじめとした実データはほとんどの場合低次元多様体に分布していると考えられており，それゆえ上記の状況は大いに起こりうる．

中間素子の degeneration をより丁寧に調べるため，hidden activation の overlap や vanishing を調べた．これは，行列 $H = (g(J_jx_i))_{ij}$ を考え，$H$ に大きく類似する列があるか，ノルムの小さな列があるかを調べれば良い．

}
}
\end{document}